\pdfoutput=1
\documentclass[11pt]{article}
\usepackage{acl}
\usepackage{times}
\usepackage{latexsym}
\usepackage[T1]{fontenc}
\usepackage[utf8]{inputenc}
\usepackage{microtype}
\usepackage{hyperref}
\usepackage{amsmath,amsfonts}
\usepackage{algorithm}
\usepackage{arydshln}
\usepackage{multirow,subfigure,booktabs}
\usepackage{caption}
\usepackage{algorithmic}
\usepackage{cite}
\usepackage{graphicx}
\usepackage{amssymb}

\usepackage{cancel}

\title{Advancing Semi-Supervised Task Oriented Dialog Systems by JSA Learning of Discrete Latent Variable Models}

\author{Yucheng Cai, Hong Liu, Zhijian Ou\thanks{~~Corresponding author.}\\
  Speech Processing and Machine Intelligence Lab\\ Tsinghua University, Beijing, China \\
  \texttt{{cyc22@mails.tsinghua.edu.cn}}\\
  \texttt{{liuhong21@mails.tsinghua.edu.cn}}\\
  \texttt{ozj@tsinghua.edu.cn}\\\And
  ~~~~~~~~~~~~~~~~Yi Huang,
  Junlan Feng \\
     ~~~~~~~~~~~~~~~~~~~~China Mobile Research Institute\\ ~~~~~~~~~~~~~~~~~~~~Beijing, China \\
  \texttt{~~~~~~~~\{huangyi,fengjunlan\}} \\
  \texttt{~~~~~~~~@chinamobile.com} \\
  }

\newcommand{\modelname}{JSA-TOD}
\newcommand{\comparemodel}{Variational}

\begin{document}

\maketitle

\begin{abstract}
Developing semi-supervised task-oriented dialog (TOD) systems by leveraging unlabeled dialog data has attracted increasing interests.
For semi-supervised learning of latent state TOD models, variational learning is often used, but suffers from the annoying high-variance of the gradients propagated through discrete latent variables and the drawback of indirectly optimizing the target log-likelihood.
Recently, an alternative algorithm, called joint stochastic approximation (JSA), has emerged for learning discrete latent variable models with impressive performances.
In this paper, we propose to apply JSA to semi-supervised learning of the latent state TOD models, which is referred to as \modelname{}.
To our knowledge, \modelname{} represents the first work in developing JSA based semi-supervised learning of discrete latent variable conditional models for such long sequential generation problems like in TOD systems.
Extensive experiments show that \modelname{} significantly outperforms its variational learning counterpart. Remarkably, semi-supervised \modelname{} using 20\% labels performs close to the full-supervised baseline on MultiWOZ2.1.

\end{abstract}

\section{Introduction}

\begin{figure}[t]
\centering
	\includegraphics[width=0.95\linewidth]{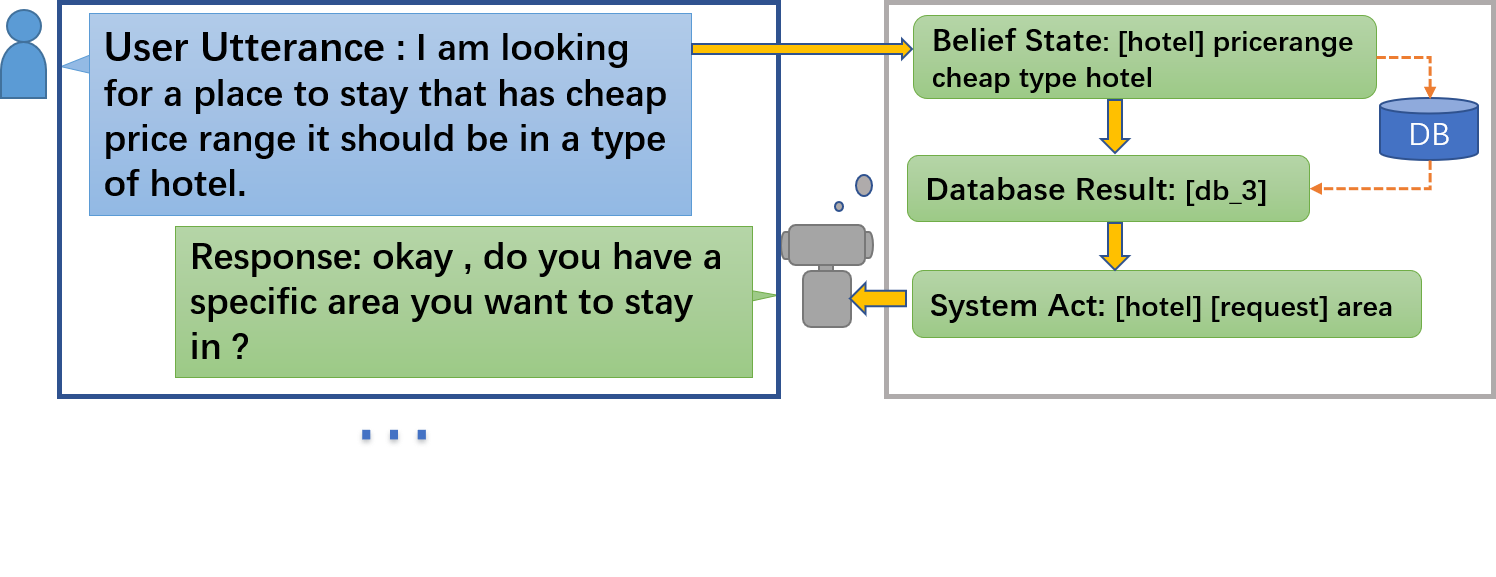}
	\vspace{-1.5em}
	\caption{The information flow in a task-oriented dialog. Square brackets denote special tokens in GPT2.}
	\vspace{-1em}
	\label{fig:flow}
\end{figure}

Task-oriented dialog (TOD) systems are designed to help users to achieve their goals through multiple turns of natural language interaction.
The system needs to parse user utterances, track dialog states, query a task-related database (DB), decide actions and generate responses, and to do these iteratively across turns.
The information flow in a task-oriented dialog is illustrated in Figure~\ref{fig:flow}.

Recent studies recast such information flow in a TOD system as conditional generation of tokens and base on pretrained language models (PLMs) such as GPT2 \citep{radford2019gpt2} and T5 \citep{raffel2020t5} as the model backbone. Fine-tuning a PLM over annotated dialog datasets such as MultiWOZ \citep{budzianowski2018large} via supervised learning has shown promising results \citep{hosseini2020simple, peng2020etal, yang2021ubar,liu2022mga}, but requires manually labeled dialog states and system acts (if used).

Notably, there are often easily-available unlabeled dialog data such as in customer-service logs and online forums. 
This has motivated the development of semi-supervised leaning (SSL) for TOD systems, which aims to leverage both labeled and unlabeled dialog data.
A broad class of SSL methods builds a latent variable model (LVM) of observations and labels and blends unsupervised and supervised learning. Unsupervised learning with a LVM usually maximizes the marginal log-likelihood, which is often intractable to compute.
Variational learning \citep{kingma2013auto} introduces an auxiliary inference model and, instead, maximizes the evidence lower bound (ELBO) of the marginal log-likelihood.
This approach of variational learning of LVMs has been studied for semi-supervised TOD systems such as in \citet{sedst, zhang-etal-2020-probabilistic,liu2021variational,li2021semi}. Particularly, discrete latent variables are mostly used, since dialog states and system acts are often modeled as taking discrete values.

However, for variational learning of discrete latent variable models, the Monte-Carlo gradient estimator for the inference model parameter is known to have high-variance. Most previous studies use the Gumbel-Softmax trick \citep{jang2016categorical} or the Straight-Through trick \citep{bengio2013estimating} empirically, which in fact are biased estimators. Another drawback of variational learning is that it indirectly optimizes the lower bound of the target marginal log-likelihood, which leaves an uncontrolled gap between the target and the bound, depending on the expressiveness of the inference model.

Recently, an alternative algorithm, called joint stochastic approximation (JSA) \citep{xu2016joint,ou2020joint}, has emerged for learning discrete latent variable models with impressive performances.
JSA directly optimizes the marginal likelihood and completely avoids gradient propagation through discrete latent variables.
In this paper, we propose to apply JSA to semi-supervised learning of the latent state TOD models, which is referred to as \modelname{}.
We develop recursive turn-level Metropolis Independence Sampling (MIS) to enable the successful application of JSA, which needs posterior sampling of the latent states from the whole dialog session. To our knowledge, \modelname{} represents the first work in developing JSA based semi-supervised learning of discrete latent variable conditional models for such long sequential generation problems like in TOD systems.

Extensive experiments show that \modelname{} significantly outperforms its variational learning counterpart in semi-supervised learning. Remarkably, semi-supervised \modelname{} using 20\% labels performs close to the supervised-only baseline using 100\% labels on MultiWOZ2.1. The code and data are released at \hyperlink{JSA-TOD}{https://github.com/cycrab/JSA-TOD}.

\section{Related Work}


\subsection{Semi-Supervised TOD Systems}
There are increasing interests in developing SSL methods for TOD systems, which aims to leverage both labeled and unlabeled data. 
Roughly speaking, there are two broad classes of SSL methods - the pretraining-and-finetuning approach and the latent variable modeling approach.
With the development of pretrained language models such as GPT2 \citep{radford2019gpt2} and T5 \citep{raffel2020t5}, the pretraining-and-finetuning approach based on backbones of PLMs has shown excellent performance for TOD systems \citep{hosseini2020simple,yang2021ubar,lee2021improving}.

Discrete latent variable models have been used for semi-supervised TOD systems \citep{sedst, zhang-etal-2020-probabilistic}\footnote{There are other previous studies of using discrete latent variable models in TOD systems, for example, \citet{wen2017latent, zhao2019rethinking,bao-etal-2020-plato}. But most of them are mainly designed to improve response generation and diversity, instead of towards semi-supervised learning. See \citet{zhang-etal-2020-probabilistic,liu2021variational} for more review of related work in latent variable models for dialogs.}, initially based on LSTM architectures. Recently, discrete latent variable models based on PLMs have been studied in \citet{liu2021variational}, combining the strengths of PLMs and LVMs for semi-supervised TOD systems. However, previous studies all resort to variational methods for learning latent variable models, which suffers from the high-variance of the gradients propagated through discrete latent variables and the drawback of indirectly optimizing the target log-likelihood.


\subsection{Joint Stochastic Approximation for Learning Latent Variable Models}
Traditionally, variational methods minimize the ``exclusive Kullback-Leibler (KL) divergence'' $KL[p||q] \triangleq \int q \log \left( \frac{q}{p} \right)$, where $p$ and $q$ are shorthands for the true posterior (of the latent variable given the observation) and its approximation (also called the inference model) respectively, in learning a latent variable model.
Recently, the JSA algorithm has been developed \citep{xu2016joint,ou2020joint}, which proposes to minimize the ``inclusive KL'' $KL[p||q] \triangleq \int p \log \left( \frac{p}{q} \right)$, which has good statistical properties that makes it more appropriate for certain inference and learning problems, particularly for those using discrete latent variables. Similar idea has been studied in a concurrent and independent work \citep{naesseth2020markovian}.
More investigations and extensions along this direction have been examined \citep{kim2020adaptive,kim2022markov}. 

In \citet{song2020jae}, JSA is applied to semi-supervised sequence-to-sequence learning, which consistently outperforms variational learning on two semantic parsing benchmark datasets. However, both generative model and inference model in \citep{song2020jae} are LSTM-based and much simpler than the ones in this work; its model complexity is similar to a single turn in a TOD system. Another difference is that this paper represents the first application of JSA in its conditional sequential version, since the latent state TOD model is a conditional sequential generative model.



\section{Preliminary: Joint Stochastic Approximation (JSA)} \label{sec:jsa}
Stochastic approximation (SA) refers to an important family of iterative stochastic optimization algorithms for stochastically solving a root finding problem, which has the form of expectations being equal to zeros \citep{robbins1951stochastic}.
Within the SA framework, the joint stochastic approximation (JSA) algorithm is recently developed \citep{xu2016joint,ou2020joint} for learning a broad class of latent variable models, particularly for learning models with discrete latent variables.
Interestingly, JSA amounts to coupling an SA version of Expectation-Maximization (SAEM) \citep{delyon1999convergence,kuhn2004coupling} with an adaptive Markov Chain Monte Carlo (MCMC) procedure.
Based on JSA, the annoying difficulty of propagating gradients through discrete latent variables and the drawback of indirectly optimizing the target log-likelihood can be gracefully addressed.




Consider a latent variable generative model $p_\theta(z, x)$ for observation $x$ and latent variable $z$, with parameter $\theta$.
Like in variational methods, JSA also jointly trains the target model $p_\theta(z,x)$ together with an auxiliary amortized inference model $q_\phi(z|x)$.
The difference is that JSA directly maximizes w.r.t. $\theta$ the marginal log-likelihood and simultaneously minimizes w.r.t. $\phi$ the inclusive KL divergence $KL(p_\theta(z|x)||q_\phi(z|x))$ between the posterior and the inference model, pooled over the training dataset:
\begin{equation}\label{eq:jsa_target}
\left\{
\begin{split}
&\min\limits_{\theta}\frac{1}{n}\sum_{i=1}^{n}\log p_\theta(x^{(i)})\\
&\min\limits_{\phi}\frac{1}{n}\sum_{i=1}^{n}KL[p_\theta(z^{(i)}|x^{(i)})||q_\phi(z^{(i)}|x^{(i)})]
\end{split}
\right.
\end{equation}
where the training dataset consists of $n$ independent and identically distributed (IID) data-points $\left\lbrace x^{(1)}, \cdots, x^{(n)} \right\rbrace$.

The optimization problem  Eq. \eqref{eq:jsa_target} can be solved by setting the gradients to zeros and applying the SA algorithm to find the root for the resulting simultaneous equations, which has the exact form of expectations equal to zeros:
\begin{align} \label{gradient}
\begin{cases}
\frac{1}{n}\sum_{i=1}^{n} E_{p_{\theta}(z^{(i)} \mid x^{(i)})}\left[\nabla_{\theta} \log p_{\theta}(x^{(i)}, z^{(i)})\right]={0} \\
\frac{1}{n}\sum_{i=1}^{n} E_{ p_{\theta}(z^{(i)} \mid x^{(i)})}\left[\nabla_{\phi} \log q_{\phi}(z^{(i)} \mid x^{(i)})\right]={0}
\end{cases}
\end{align}

The resulting JSA algorithm, as summarized in Algorithm \ref{alg:JAE}, iterates Monte Carlo sampling and parameter updating.
In each iteration, we draw a training observation $x^{(\kappa)}$ and then sample $z^{(\kappa)}$ through Metropolis Independence Sampling (MIS),
with $p_\theta(z^{(\kappa)}|x^{(\kappa)})$ as the target distribution and $q_\phi(z|x^{(\kappa)})$ as the proposal: 

1) Propose $z \sim q_\phi(z|x^{(\kappa)})$;

2) Accept $z^{(\kappa)}=z$ with probability
\begin{displaymath}
min\left\lbrace 1, \frac{w(z)}{w(\bar{z}^{(\kappa)})} \right\rbrace
\end{displaymath}
where $w(z) = \frac{p_\theta(z|x^{(\kappa)})}{q_\phi(z|x^{(\kappa)})} \propto \frac{p_\theta(z,x^{(\kappa)})}{q_\phi(z|x^{(\kappa)})}$ is the usual importance ratio between the target and the proposal distribution and $\bar{z}^{(\kappa)}$ denotes the cached latent state for observation $x^{(\kappa)}$.

\begin{algorithm}[tb]
	\caption{The JSA algorithm}
	\label{alg:JAE}
	\begin{algorithmic}
		\REPEAT
		\STATE \underline{Monte Carlo sampling:}\\
		Draw $\kappa$ over $1,\cdots,n$, pick the data-point $x^{(\kappa)}$ along with the cached $\bar{z}^{(\kappa)}$, and use MIS to draw $z^{(\kappa)}$;
		\STATE \underline{Parameter updating:}\\
		Update $\theta$ by ascending: $\nabla_\theta \log p_\theta(z^{(\kappa)},x^{(\kappa)})$;\\
		Update $\phi$ by ascending: $\nabla_\phi \log q_\phi(z^{(\kappa)}|x^{(\kappa)})$;
		\UNTIL{convergence}
	\end{algorithmic}
\end{algorithm}

The JSA algorithm can be intuitively understood as a stochastic extension of the well-known EM algorithm \citep{Dempster1977MaximumLF}.
Since the latent variable $z^{(\kappa)}$ is unknown for data-point $x^{(\kappa)}$, the Monte Carlo sampling step in JSA fills the missing value for $z^{(\kappa)}$ through sampling $p_\theta(z^{(\kappa)}|x^{(\kappa)})$, which is analogous to the E-step in EM.
Then in the parameter updating step, $z^{(\kappa)}$ is treated as if being known, and used to optimize over $\theta$ and $\phi$ by performing gradient ascent using $\nabla_\theta \log p_\theta(z^{(\kappa)},x^{(\kappa)})$ and $\nabla_\phi \log q_\phi(z^{(\kappa)}|x^{(\kappa)})$ respectively. This is analogous to the M-step in EM, but with the proposal $q_\phi$ being adapted as well.
In summary, we could refer to the underlying mechanism of JSA as Propose, Accept/Reject, and Optimize (or, for short, the PARO mechanism), which establishes JSA as a simple, solid and effective approach to learning discrete latent variable models.


\section{Method}

\begin{figure}[t]
\centering
	\includegraphics[width=1\linewidth]{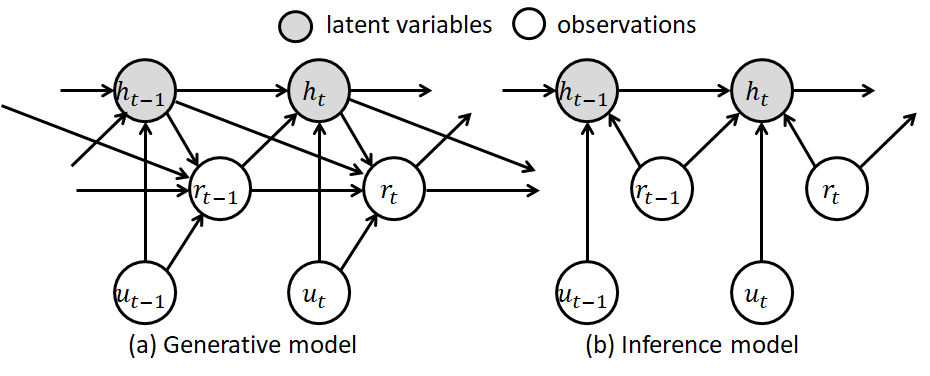}
 	\vspace{-1em}
	\caption{The probabilistic graphical model of Markov latent state generative model (a) and inference model (b) for TOD systems. $u_t$ and $r_t$ are user utterance and system response respectively. The latent variables $h_t=\{b_t, a_t\}$ are the concatenation of dialog state and system act, which specifically are represented by token sequences in our experiments.}
 	\vspace{-0.5em}
	\label{fig:overview}
\end{figure}

\subsection{Definition of Discrete Latent Variables in TOD systems}
In a TOD system, let $u_t$ denote the user utterance, $b_t$ the dialog state, $db_t$ the DB result, $a_t$ the system act and $r_t$ the delexicalized response, respectively, at turn $t$. In this work, all these variables are converted to token sequences, like in DAMD \citep{zhang2020task}.
As shown in Figure \ref{fig:flow}, the workflow for a TOD system is, for each dialog turn $t$, to generate $b_t$, $a_t$ and $r_t$, given $u_t$ and dialog history $u_1, r_1, \cdots, u_{t-1}, r_{t-1}$. The database result $db_t$ is deterministically obtained by querying database using the predicted $b_t$, and thus could be omitted in the following probabilistic modeling of a TOD system for simplicity.

Let $h_t=\{b_t, a_t\}$
denote the concatenation of dialog state and system act.
Specifically, dialog state $b_t$ and system act $a_t$ are represented by sequences of labels, for example, $[train] \ day\ monday\ [hotel]\ pricerange\  cheap$ and $[train]\ [inform]\ choice\ departure\ [request]$ $ destination$, respectively.
Notably, $h_t$'s are observed in labeled dialogs, but they become latent variables in unlabeled dialogs in training and need to be generated in testing.
With this definition of $h_t$'s, latent variable models can be developed for TOD systems, which will be described shortly in the next subsection.

Remarkably, the above definition of latent variables as sequences of labels in this paper is similar to \citet{zhang-etal-2020-probabilistic,liu2021variational}. An important feature of such latent variables is that they are sensible and interpretable, which correspond to meaningful annotations according to the task knowledge. It is only in unlabeled dialogs that they become unobservable.
This is different in nature from some other previous studies of using latent variables in TOD models \citep{wen2017latent, zhao2019rethinking,bao-etal-2020-plato}, where the latent variables are just assumed to be $K$-way categorical variables and learned in a purely data driven way.

\subsection{A Probabilistic Latent State TOD Model}
With the above introduction of latent variables and motivated by recent studies \citep{zhang-etal-2020-probabilistic,liu2022mga}, the workflow of a TOD system could be described by a conditional sequential generative model with latent variables $h_t$'s as follows for $T$ turns, with parameter $\theta$:
\begin{align}
&p_\theta(h_{1:T}, r_{1:T}|u_{1:T}) \nonumber\\
=&\prod_{t=1}^T p_\theta(h_{t},r_{t}|u_1, h_1, r_1, \cdots, u_{t-1}, h_{t-1}, r_{t-1}, u_{t}) \label{eq:LS-1}\\
=&\prod_{t=1}^T p_\theta(h_{t},r_{t}|h_{t-1}, r_{t-1}, u_{t})\text{(by Markov assumption)} \label{eq:LS-2}
\vspace{-1em}
\end{align}
Here Eq. \eqref{eq:LS-1} and Eq. \eqref{eq:LS-2} could be collectively referred to as \emph{latent state TOD models}, being non-Markov and Markov respectively.
Eq. \eqref{eq:LS-1} represents non-Markov latent state models, which, with different further instantiations, are used in recent PLM-based TOD systems such as in \citet{hosseini2020simple,yang2021ubar,liu2021variational}.
In contrast, Eq. \eqref{eq:LS-2} makes the Markov assumption that the conditional generation of current $h_t$ and $r_t$ (when given $u_t$) depends on the dialog history only through $h_{t-1}$ and $r_{t-1}$ at the immediately preceding turn.
Markov models have been employed in LSTM-based TOD systems such as in \citet{lei2018sequicity,zhang2020task,zhang-etal-2020-probabilistic}.
A recent study in \citet{liu2022mga} revisits Markovian generative architectures (MGAs) for PLM backbones (GPT2 and T5) and shows their efficiency advantages in memory, computation and learning over non-Markov models.

\subsection{Model Instantiation and Supervised Learning} \label{sec:model-instantiation}
In our experiments, we mainly consider MGA based latent state TOD systems \citep{liu2022mga}, which are illustrated in Figure \ref{fig:overview} as directed probabilistic graphical models. 
The conditional distribution $p_\theta(h_{t},r_{t}|h_{t-1}, r_{t-1}, u_{t})$ is instantiated as
\begin{equation} \label{eq:turn-p}
p_\theta(b_{t},a_{t},r_{t}|b_{t-1}, r_{t-1}, u_{t})
\end{equation}
which is realized based on a GPT2 backbone in our experiments.
The concatenation $b_{t-1} \oplus r_{t-1} \oplus u_{t}$ is used as the conditioning input, and the output $b_{t} \oplus a_{t} \oplus r_{t}$ is generated token-by-token in an auto-regressive manner, where $\oplus$ denotes the concatenation of token sequences.

In order to perform unsupervised learning over unlabeled dialogs (to be detailed below), we introduce an inference model $q_\phi(h_{1:T}|u_{1:T},r_{1:T})$ as follows to approximate the true posterior $p_\theta(h_{1:T}|u_{1:T},r_{1:T})$:
\begin{equation} \label{eq:tod-q}
q_\phi(h_{1:T}|u_{1:T},r_{1:T})=\prod_{t=1}^{T} q_\phi(h_{t}|h_{t-1},r_{t-1},u_t,r_t)
\end{equation}
The conditional $q_\phi(h_{t}|h_{t-1},r_{t-1},u_t,r_t)$ is instantiated as
\begin{equation} \label{eq:turn-q}
q_\phi(b_{t},a_t|b_{t-1},r_{t-1},u_t,r_t)
\end{equation}
which is realized based on a GPT2 backbone as well in our experiments.

When labeled dialog data are available, the supervised training of the latent state generative model $p_\theta$ in and inference model $q_\phi$ can be decomposed into turn-level teacher-forcing, since the latent states $h_t$'s are known (labeled) for all turns and the model likelihoods decomposes over turns, as shown in Eq. \eqref{eq:LS-2} and Eq. \eqref{eq:tod-q}.

\subsection{JSA Learning over Unlabeled Dialogs}
Suppose that we have $n$ unlabeled dialogs $\left\lbrace (u_{1:T_i}^{(i)},r_{1:T_i}^{(i)})|i=1,\cdots,n\right\rbrace$, i.e., user utterances and system responses are available for each dialog, but without any annotations of the latent states. The training instances are indexed by the superscripts, and $T_i$ denote the number of turns in the $i$-th training instance.
The unsupervised learning of the latent state TOD model over such unlabeled data can be realized by applying the JSA algorithm, and more specifically its conditional version, to maximize the conditional marginal log-likelihood $\log p_\theta(r_{1:T}|u_{1:T})$.


The objective functions in JSA learning can be developed as follows, similar to Eq. \eqref{eq:jsa_target}:
\begin{displaymath}
\left\{
\begin{split}
\min \limits_{\theta}\frac{1}{n}\sum_{i=1}^{n}\log p_\theta(r_{1:T_i}^{(i)}|&u_{1:T_i}^{(i)}) \\
\min \limits_{\phi}\frac{1}{n}\sum_{i=1}^{n}KL[  p_\theta(h_{1:T_i}^{(i)}|&u_{1:T_i}^{(i)},r_{1:T_i}^{(i)})\\
&||q_\phi(h_{1:T_i}^{(i)}|u_{1:T_i}^{(i)},r_{1:T_i}^{(i)})]\\
\end{split}
\right.
\end{displaymath}
where we substitute observation $x$ by $r_{1:T}$ and latent variable $z$ by $h_{1:T}$, all conditioned on $u_{1:T}$.

Basically, JSA learning iterates Monte Carlo sampling and parameter updating, as outlined in Algorithm \ref{alg:JAE}.
In each iteration, we randomly pick a training instance $(u_{1:T}, r_{1:T})$ along with the cached latent state $\bar{h}_{1:T}$, and we need to draw a posterior sample $h_{1:T} \sim p_\theta(h_{1:T}|u_{1:T},r_{1:T})$.
Remarkably, it can be shown in Appendix \ref{sec:proof-post-recur} that for the posterior $p_\theta(h_{1:t}|u_{1:t},r_{1:t})$ induced from the joint distribution in Eq. \eqref{eq:LS-2}, the following recursion holds:
\begin{equation} \label{eq:post-recur}
\begin{split}
&p_\theta(h_{1:t}|u_{1:t},r_{1:t})\\
\propto &p_\theta(h_{1:t-1}|u_{1:t-1},r_{1:t-1})p_\theta(h_t,r_t|h_{t-1}, r_{t-1},u_t)
\end{split}
\end{equation}
Based on such recursion, we can develop a recursive turn-level MIS sampler, as shown in Algorithm \ref{alg:sampling}, which recursively runs MIS sampler turn-by-turn and finally obtains a valid posterior sample for the whole dialog session, i.e., $h_{1:T} \sim p_\theta(h_{1:T}|u_{1:T},r_{1:T})$.

Suppose that we have obtained a sample for the previous $t$-1 turns, i.e., $h_{1:t-1} \sim p_\theta(h_{1:t-1}|u_{1:t-1},r_{1:t-1})$. Then, we perform MIS sampling as follows, with $p_\theta(h_{1:t}|u_{1:t},r_{1:t})$ as the target distribution and 
\begin{equation} \label{eq:mis-porposal}
p_\theta(h_{1:t-1}|u_{1:t-1},r_{1:t-1}) q_\phi(h_{t}|h_{t-1},r_{t-1},u_t,r_t)
\end{equation}
as the proposal distribution:

1) Propose $h'_{t} \sim q_\phi(h_{t}|h_{t-1},r_{t-1},u_t,r_t)$. Thus, 
$(h_{1:t-1}, h'_{t})$ is a valid sample proposed from the proposal distribution as shown in Eq. \eqref{eq:mis-porposal};

2) Simulate $\xi \sim Uniform[0,1]$ and let
\begin{equation} \label{eq:tod-accept-reject}
h_{t} = \left\{
\begin{split}
&h'_{t}, &&\text{if}~\xi \le min\left\lbrace 1, \frac{w(h_{1:t-1}, h'_{t})}{w(h_{1:t-1}, \bar{h}_{t})} \right\rbrace \\
&\bar{h}_t, &&\text{otherwise}
\end{split}
\right.
\end{equation}
where the importance ratio between the target and the proposal distribution
\begin{align}
&w(h_{1:t-1}, h_{t}) \nonumber \\
=&\frac{p_\theta(h_{1:t}|u_{1:t},r_{1:t})}{p_\theta(h_{1:t-1}|u_{1:t-1},r_{1:t-1}) q_\phi(h_{t}|h_{t-1},r_{t-1},u_t,r_t)} \nonumber \\ 
\propto & \frac{p_\theta(h_t,r_t|h_{t-1}, r_{t-1},u_t)}{q_\phi(h_{t}|h_{t-1},r_{t-1},u_t,r_t)}  \label{eq:tod-MIS-weight}
\end{align}

\begin{algorithm}[tb]
  \caption{Recursive turn-level MIS sampler}
  \label{alg:sampling}
  \begin{algorithmic}
  \REQUIRE A $T$-turn dialog $(u_{1:T},r_{1:T})$ with cached latent state $\bar{h}_{1:T}$, generative model $p_\theta$ in Eq. \eqref{eq:LS-2}, inference model $q_\phi$ in Eq. \eqref{eq:tod-q}.
    \FOR{$t=1$ to $T$}
    \STATE Propose $h'_{t} \sim q_\phi(h_{t}|h_{t-1},r_{t-1},u_t,r_t)$;
    \STATE Accept $h'_{t}$ as $h_t$, or reject $h'_{t}$ and keep $\bar{h}_{t}$ as $h_t$, according to Eq. \eqref{eq:tod-accept-reject};
    \ENDFOR
  \STATE {\bf Return:} $h_{1:T}$, as a posterior sample from $p_\theta(h_{1:T}|u_{1:T},r_{1:T})$ and used as the new cached latent state.
  \end{algorithmic}
\end{algorithm}


After we obtain the sampled latent state $h_{1:T}$ from Algorithm \ref{alg:sampling}, we perform parameter updating, as outlined in Algorithm \ref{alg:JAE}. 
The sampled latent state $h_{1:T}$ is treated as if being known, and we can calculate the gradients of $\log p_\theta(h_{1:T}, r_{1:T}|u_{1:T})$ and  $\log q_\phi(h_{1:T}|u_{1:T},r_{1:T})$  w.r.t. $\theta$ and $\phi$ according to Eq. \eqref{eq:LS-2} and Eq. \eqref{eq:tod-q} respectively, as if we calculate gradients in supervised training.
Thanks to the PARO mechanism of JSA, we have such a conceptual simplicity for learning seemly complex conditional sequential latent variable model.


\subsection{Semi-Supervised TOD Systems via JSA}

Now we have introduced the method of building latent state TOD systems (Eq. \eqref{eq:LS-2} and Eq. \eqref{eq:tod-q}) with JSA learning (Algorithm \ref{alg:JAE} and Algorithm \ref{alg:sampling}), which is referred to \modelname{}.
Semi-supervised learning over a mix of labeled and unlabeled data could be readily realized in \modelname{} by maximizing the weighted sum of $\log p_\theta(h_{1:T}, r_{1:T}|u_{1:T})$ (the conditional joint log-likelihood) over labeled data and $\log p_\theta(r_{1:T}|u_{1:T})$ (the conditional marginal log-likelihood) over unlabeled data.

The semi-supervised training procedure of \modelname{} is summarized in Algorithm \ref{alg:training}.
Specifically, we first conduct supervised pre-training of both the generative model $p_\theta$ and the inference model $q_\phi$ on labeled data in \modelname{}. 
Then we randomly draw supervised and unsupervised mini-batches from labeled and unlabeled data. For labeled dialogs, the latent states $h_t$'s are given (labeled). For unlabeled dialogs, we apply the recursive turn-level MIS sampler (Algorithm \ref{alg:sampling}) to sample the latent states $h_t$'s\footnote{Sampling is empirically implemented via greedy decoding in our experiments.} and treat them as if being given.
The gradients calculation and parameter updating are then the same for labeled and unlabeled dialogs.
Such simplicity in application is an appealing property of JSA, apart from its superior performance, as we show later in experiments.

\begin{algorithm}[tb]
  \caption{Semi-supervised training in \modelname{}}
  \label{alg:training}
  \begin{algorithmic}
        \REQUIRE A mix of labeled and unlabeled dialogs.
	    \STATE Run supervised pre-training of $\theta$ and $\phi$ on labeled dialogs;
	    \REPEAT
	    \STATE Draw a dialog $(u_{1:T},r_{1:T})$;
		\IF  {$(u_{1:T},r_{1:T})$ is not labeled}
		\STATE Generate $h_{1:T}$ by applying the recursive turn-level MIS sampler (Algorithm \ref{alg:sampling}); 
		\ENDIF
		\STATE $J_\theta=0, J_\phi=0$;
        \FOR {$i = 1,\cdots,T$}
        \STATE $J_\theta += \log p_\theta(h_{t},r_{t}|h_{t-1}, r_{t-1}, u_{t})$;
        \STATE $J_\phi += \log q_\phi(h_{t}|h_{t-1},r_{t-1},u_t,r_t)$;
        \ENDFOR
        \STATE Update $\theta$ by ascending: $\nabla_\theta J_\theta$;
        \STATE Update $\phi$ by ascending: $\nabla_\phi J_\phi$;
        \UNTIL {convergence}
        \RETURN {$\theta$ and $\phi$}
  \end{algorithmic}
\end{algorithm}

\section{Experiments}
\subsection{Experiment settings} 
Experiments are conducted on MultiWOZ2.1 \citep{eric2019multiwoz}, which is an English multi-domain dialogue dataset of human-human conversations, collected in a Wizard-of-Oz setup with 10.4k dialogs over 7 domains.
The dataset was officially randomly split into a train, test and development set, which consist of 8434, 1000 and 1000 dialog samples, respectively.
The dialogs in the dataset are all labeled with dialog states and system acts at every turn.
Compared to MultiWOZ2.0, MultiWOZ2.1 removed some noisy state values.
Following \citep{liu2022mga}, some inappropriate state values and spelling errors are further corrected.
Dialog responses are delexicalized to reduce surface language variability. We implement domain-adaptive pre-processing like in DAMD \citep{zhang2020task}.
More implementation details for our experiments are available in Appendix \ref{sec:implementation}. 

For evaluation in MultiWOZ2.1, there are mainly four metrics for corpus based evaluation \citep{mehri2019structured}.
\emph{Inform Rate} measures how often the entities provided by the system are correct; \emph{Success Rate} refers to how often the system is able to answer all the requested attributes by user; \emph{BLEU Score} is used to measure the fluency of the generated responses by analyzing the amount of n-gram overlap between the real responses and the generated responses; \emph{Combined Score} is computed as (BLEU + 0.5 * (Inform + Success)). 
To avoid any inconsistencies in evaluation, we use the evaluation scripts in \citet{nekvinda2021shades}, which are now also the standardized scripts adopted in the MultiWOZ website.

\subsection{Main Results}

\begin{table}[t]
    \caption{Main results on MultiWOZ2.1 for comparison between supervised-only, variational, and JSA methods.
    Results are reported as the mean and standard deviation from 3 runs with different random seeds.}
	\centering
	\resizebox{1\linewidth}{!}{
			\begin{tabular}{cc cccc cccc}
				\toprule
				Proportion &Method  &Inform &Success &BLEU &Combined \\ 
				\midrule
				100\% &Sup-only  &84.50$\pm$0.29  &72.77$\pm$0.50  &18.96$\pm$0.36  &97.59$\pm$0.54\\
				\midrule
				\multirow{3}{*}{20\%} 
				&Sup-only   & 75.70$\pm$1.87  & 61.07$\pm$2.21  & 16.66$\pm$0.29  & 85.05$\pm$2.16 \\
				&\comparemodel{}   & 81.83$\pm$1.55  & 67.67$\pm$0.50  & 17.88$\pm$0.95  & 92.63$\pm$0.30 \\
				& JSA & 83.25$\pm$ 0.65 & 71.40$\pm$1.20  & 18.72$\pm$0.07  & \textbf{96.04$\pm$0.85} \\
				\midrule
				\multirow{3}{*}{15\%} 
				&Sup-only   & 80.00$\pm$0.43  & 55.57$\pm$1.22  & 16.20$\pm$0.24  & 79.00$\pm$1.51 \\
				&\comparemodel{}  & 80.85$\pm$0.65  & 67.67$\pm$0.88  & 17.68$\pm$0.29  & 91.86$\pm$0.19 \\
				&JSA  & 83.23$\pm$0.53 & 71.97$\pm$1.27  & 18.59$\pm$0.19  & \textbf{95.47$\pm$0.73} \\
				\midrule
				\multirow{3}{*}{10\%} 
				&Sup-only   & 67.57$\pm$0.39  & 50.03$\pm$1.09  & 15.31$\pm$0.28  & 74.11$\pm$0.59 \\
				&\comparemodel{}   & 80.67$\pm$1.33  & 66.97$\pm$1.23  & 17.34$\pm$0.56  & 91.15$\pm$1.80 \\
				&JSA  & 81.97$\pm$0.79 & 70.40$\pm$0.99  & 18.09$\pm$0.38 & \textbf{94.27$\pm$1.23} \\
				\midrule
				\multirow{3}{*}{5\%} 
				&Sup-only   & 49.73$\pm$2.45  & 33.67$\pm$1.79  & 14.07$\pm$0.13  & 55.77$\pm$1.94 \\
				&\comparemodel{}   & 74.17$\pm$0.53 & 59.93$\pm$0.34  & 16.06$\pm$0.69  & 83.11$\pm$1.04 \\
				&JSA  & 72.37$\pm$1.19 &59.73$\pm$0.92 & 18.57$\pm$0.54 & \textbf{84.62$\pm$0.43} \\
				\bottomrule
		\end{tabular}}
	\vspace{-0.5em}
		\label{compare-results}
\end{table}

In the semi-supervised experiments, we randomly draw some proportions (5\%, 10\%, 15\% and 20\%) of the labeled dialogs from the MultiWOZ2.1 training set, with the rest dialogs in the training set treated as unlabeled, and conduct semi-supervised experiments.
Specifically, the number of dialogs kept as labeled under these proportions are 1686, 1265, 843, and 421, respectively, while the rest dialogs are used as unlabeled (i.e., the original labels of dialog states and system acts at all turns are removed for those dialogs in the training set).

The main results are shown in Table \ref{compare-results}.
For model instantiations, we use the GPT2 based Markov generative model and inference model, as introduced in \citet{liu2022mga}.
It has been shown in \citet{liu2022mga} that using Markovian generative architecture achieves better results than non-Markov models in the low-resource setting for both supervised-only learning and semi-supervised variational learning, which makes it a strong baseline to compare.
We first train the generative model and inference model on only the labeled data, which is referred to as ``Supervised-only'' (Sup-only for short). 
Then, we perform semi-supervised training on both labeled and unlabeled data.
Using the variational method in \citep{liu2021variational,liu2022mga}, we get the baseline results of ``\comparemodel{}'', where 
the Straight-Through trick is used to propagate the gradients through discrete latent variables.
Using the JSA method proposed in Algorithm \ref{alg:training}, we get the results of ``JSA''. 
We conduct the experiments with 3 random seeds and report the mean and standard deviation in Table \ref{compare-results}. 

From Table \ref{compare-results}, we can see that both the Variational and the JSA methods outperform the Supervised-only method substantially across all label proportions. This clearly demonstrate the advantage of semi-supervised TOD systems. Remarkably, semi-supervised \modelname{} using 20\% labels performs close to the supervised-only baseline using 100\% labels on MultiWOZ2.1.

When comparing the two semi-supervised methods, JSA performs better than Variational significantly across almost all label proportions in terms of all four metrics (Inform Rate, Success Rate, BLEU, and Combined Score). 
Exceptionally, in the case of 5\% labels, the Inform Rate of JSA is worse than that of Variational, the Success Rates are close; Nevertheless, the Combined Score of JSA is significantly better. Presumably, this is because we use the Combined Scores to monitor the training, apply early stopping and select the model with the best Combined Score on the validation set. Such model selection put more priority on the overall performance in terms of Combined Scores.

Further, the results in Table \ref{compare-results} are pooled over all label proportions and all random seeds, 
and the matched-pairs significance tests \citep{gillick1989some} are conducted to compare JSA and Variational for Inform, Success and BLEU respectively.
The p-values are $9.27\times10^{-2}$, $2.576\times10^{-14}$, and $2.939\times10^{-39}$ respectively, which show that JSA significantly outperforms Variational.





\subsection{Ablation study and analysis}

Notably, the JSA and the variational methods in our experiments use the same model instantiations for $p_\theta$ and $q_\phi$. The only difference lies in the learning methods they used. In the following, we provide ablation study to illustrate the superiority of JSA over variational in learning latent state TOD models.

\textbf{The importance of Metropolis Independence Sampling in JSA.~}In JSA, we need to use Monte Carlo sampling, particularly the Metropolis Independence Sampling (MIS) to decide whether or not to update the cached latent states $h_t$'s. A naive method is to always accept the labels proposed by the inference model, which is somewhat like self-training \citep{rosenberg2005semi}. 
Another simple method is to run session-level MIS, with the whole Eq. \eqref{eq:LS-2} as the target distribution and the whole Eq. \eqref{eq:tod-q} as the proposal distribution. The whole $h_{1:T}$ is proposed via ancestral sampling and then get accepted/rejected.
The results from one run with each different method are shown in Table \ref{MIS-compare}. 
Both MIS based methods significantly improves the results, which clearly reveals the importance of using MIS in JSA.
By the accept/reject mechanism, we accept latent states which have higher importance ratios and exploit them to update both generative model and inference model, and at the same time, we also explore the state space by randomly accepting latent states which have lower importance ratios.
Exploitation and exploration of the latent states seems to be well balanced in JSA, which may explain its good performance.
Our proposed recursive turn-level MIS in Algorithm \ref{alg:sampling} clearly outperforms the session-level MIS, since it samples in a much lower dimensional state space.

	\begin{table}[t]
		\caption{Ablation results for using different methods to update latent states $h_t$'s (label proportion: 10\%, random seed:11)
		}
		\vspace{-0.5em}
		\centering
		\resizebox{\linewidth}{!}{
			\begin{tabular}{cc cccc}
				\toprule
				Method  &Inform &Success &BLEU &Combined \\
				\midrule
				Without MIS  &71.10 &59.80 &18.71 &84.16\\
				Session-level MIS  &78.70     &65.50 &17.20 &89.30\\
				Recursive turn-level MIS  &82.80   
 & 71.80  & 18.56 & \textbf{95.86}\\
         \bottomrule	
		\end{tabular}}
		\vspace{-1.0em}
		\label{MIS-compare}
	\end{table}

\textbf{The latent state prediction performance of inference model.~}In both variational and JSA learning, the inference model $q_\phi$, which is introduced to approximate the true posterior, plays an important role. 
The latent states inferred from $q_\phi$ are used, either directly as in variational learning or after accepted/rejected as in JSA learning, to optimize the generative model $p_\theta$.
We measure the quality of the latent states predicted from $q_\phi$ by label precision/recall/F1, compared to oracle $b_t$ and $a_t$ (excluding $db_t$)
. We compare different $q_\phi$ obtained from three training methods - Supervised-only, Variational, and JSA. Note that at the end of running any particular training method, we obtain not only $p_\theta$ but also $q_\phi$. The performances of $p_\theta$ over the test set are shown in Table \ref{compare-results}.
The testing performances of $q_\phi$ obtained from one run of each different method are shown in Table \ref{posterior-accuracy}.
It can be seen that semi-supervised variational learning does not improve the prediction ability of the inference model, compared to the inference model trained only on the labeled data. In contrast, the prediction performance of the inference model is increased significantly by semi-supervised JSA learning, which is in line with the superior results of JSA's generative model as shown in Table \ref{compare-results}.

    \begin{table}[t]
		\caption{Performance comparison of inference models from different methods, measured by latent state prediction precision/recall/F1 over the test set.}
		\vspace{-0.5em}
		\centering
		\resizebox{\linewidth}{!}{
			\begin{tabular}{cc cccc}
				\toprule
				Label Proportion  &Method &Precision &Recall &F1 \\
				\midrule
				\multirow{3}{*}{20\%} 
				&Supervised-only  &0.928 & 0.908 &  0.918  \\
				&\comparemodel{}  & 0.924 & 0.900 & 0.912 \\
				&JSA  & \textbf{0.936} & \textbf{0.925} & \textbf{0.931} \\
				\midrule
				\multirow{3}{*}{15\%} 
				&Supervised-only  &0.924& 0.891&0.907 \\
				&\comparemodel{}  & 0.917 & 0.872 & 0.894   \\
				&JSA  & \textbf{0.934} & \textbf{0.910} & \textbf{0.922} \\
				\midrule
				\multirow{3}{*}{10\%} 
				&Supervised-only  & 0.916& 0.868 & 0.891  \\
				&\comparemodel{}  & 0.887 & 0.880 &0.883   \\
				&JSA  & \textbf{0.930} & \textbf{0.898} & \textbf{0.914} \\
				\midrule
				\multirow{3}{*}{5\%} 
				&Supervised-only  &0.894& 0.804&0.847  \\
				&\comparemodel{}  & 0.891 & 0.838 &0.864   \\
				&JSA  & \textbf{0.904} & \textbf{0.863} & \textbf{0.883} \\
         \bottomrule	
		\end{tabular}}
		\label{posterior-accuracy}
	\end{table}
	
\begin{figure}[t]
\centering
	\includegraphics[width=0.95\linewidth]{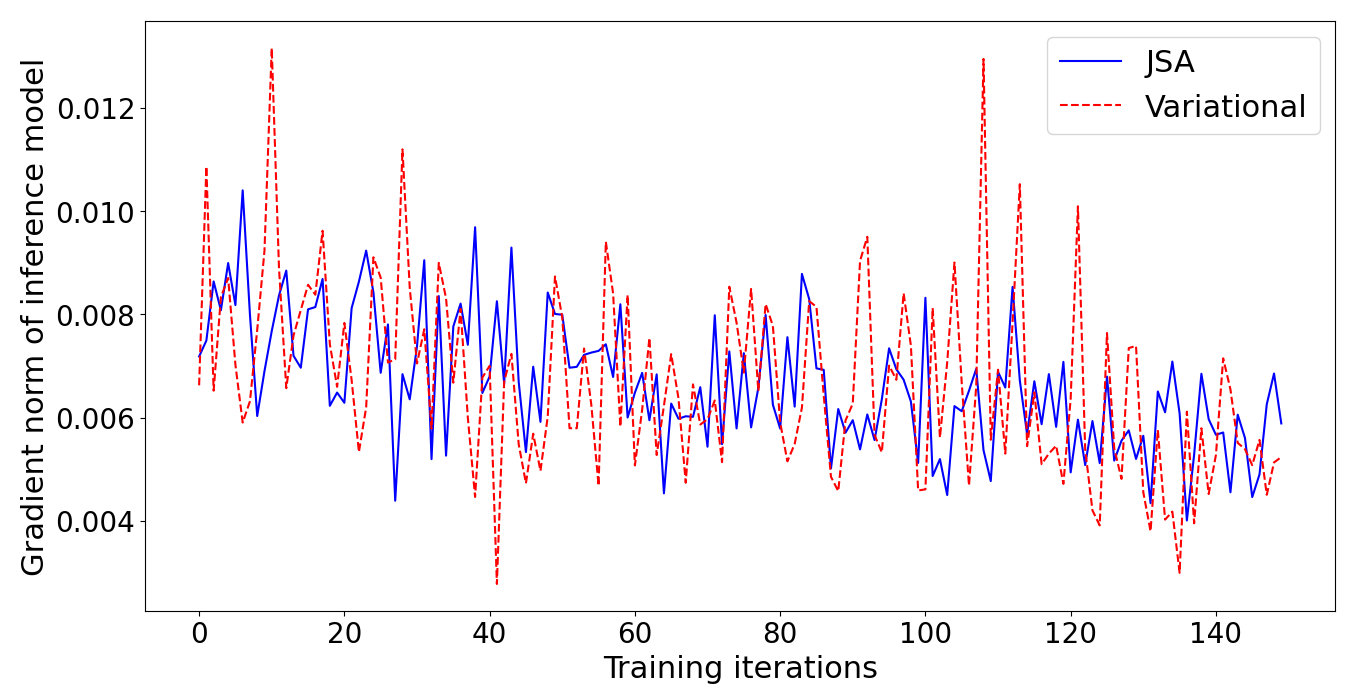}
	\vspace{-0.5em}
	\caption{Comparison of the gradient norms from the inference models during training, using variational and JSA methods respectively (label proportion: 10\%).}
	\vspace{-0.5em}
	\label{fig:variance}
\end{figure}

\textbf{The variance of the gradients from inference model.~}
The gradients for the inference model parameters in variational learning are known to have high-variance, due to gradient propagation through discrete latent variables, while JSA avoids such drawback.
From one run of semi-supervised learning under 10\% labels, we plot the gradient norms for the inference model parameters, from using the variational and the JSA methods respectively, which are shown in Figure \ref{fig:variance}. For clarity of comparison, we normalize the sum of the gradient norms over all iterations to be one. 
It can be clearly seen from Figure \ref{fig:variance} that the gradients during variational training are more noisy than those in JSA tranining.
Specifically, the variances of the time-series of the gradient norms in Figure \ref{fig:variance} are $3.097 \times 10^{-6}$ and $1.527 \times 10^{-6}$ for the variational  and the JSA methods respectively.

\textbf{Pretrained models on external dialog corpora can be further improved by JSA learning for semi-supervised TOD systems.~}Pretraining and LVM based learning are two broad classes of semi-supervised methods.
Recently, pretraining on external dialog copora has also shown to be promising for building TOD systems for low-resource scenarios \citep{peng2020etal,su2021multitask}. In this section, we show that JSA learning can be used to further improve over such pretrained models.
We use four dialog corpora - MSRE2E \citep{li2018microsoft},  Frames \citep{el2017frames}, TaskMaster \citep{byrne2019taskmaster} and SchemaGuided \citep{rastogi2020towards}, which consist of 16545 dialogs with human annotations on belief states and dialog acts, and we follow the preprocessing in \citep{su2021multitask}.
Generative model and inference model, initialized from GPT2, are pretrained separately on those four corpora, the same as that in supervised-pretraining. 
Then, we conduct semi-supervised training with only 3\% labels in MultiWOZ2.1 (i.e., 240 labeled dialogs with the rest being unlabeled).
The results in Table \ref{pretrain-boost} show that semi-supervisded JSA on top of pretrained models obtains the best result. This is an encouraging result from using 3\% labels, which is close to the naive supervised-only method using 20\% labels as shown in Table \ref{compare-results}.

    \begin{table}[t]
		\caption{MultiWOZ2.1 testing results for different methods (label proportion: 3\%). ``+Pretrained model'' means that the method is initialized from the pretrained models over four external dialog corpra.
		}
 		\vspace{-0.5em}
		\centering
		\resizebox{\linewidth}{!}{
			\begin{tabular}{lcccc}
				\toprule
				Method  &Inform &Success &BLEU &Combined \\
				\midrule
				Supervised-only  &38.70    &25.60  &16.42  &48.57  \\
				~+ Pretrained model  &57.70  &39.30  &14.15  &62.65 \\
				Semi-supervised JSA   &55.50  &44.80  &16.93  &67.08   \\
				~+ Pretrained model  &\textbf{73.00}  &\textbf{58.60}  &\textbf{18.61}  &  \textbf{84.41}  \\
         \bottomrule	
		\end{tabular}}
		\label{pretrain-boost}
	\end{table}
	
\section{Conclusion and Future Work}
This paper represents a progress towards building semi-supervised TOD systems by learning latent state TOD models.
Traditionally, variational learning is often used; notably, the recently emerged JSA method has been shown to surpass variational learning, particularly in learning of discrete latent variable models.
This paper represents the first application of JSA in its conditional sequential version, particularly for such long sequential generational problems like in TOD systems.
Extensive experiments clearly show the superiority of \modelname{} over its variational learning counterpart, not only in benchmark metrics for semi-supervised TOD systems but also from the latent state prediction performances and the variances of the gradients of the inference model.
Since discrete latent variable models are widely used in many natural language procession tasks, we hope the results presented in this paper will encourage the community to further explore the applications of JSA and improve upon current approaches.

\section{Acknowledgements}
This research was funded by Joint Institute of Tsinghua University - China Mobile Communications Group Co., Ltd., Beijing, China.

\bibliography{JSA}

\clearpage
\appendix

\section{Proof of Eq. \eqref{eq:post-recur}}
\label{sec:proof-post-recur}
First, we have
\begin{align*}
&p_\theta(h_{1:t},r_{1:t}|u_{1:t})\\
=&p_\theta(h_{1:t-1},r_{1:t-1}|u_{1:t-1}, u_t)\\
& \times p_\theta(h_t,r_t|u_{1:t-1}, u_t, h_{1:t-1}, r_{1:t-1})\\
=&p_\theta(h_{1:t-1},r_{1:t-1}|u_{1:t-1}, \cancel{u_t})\\
& \times p_\theta(h_t,r_t|h_{t-1}, r_{t-1}, u_t, \cancel{h_{1:t-2}, r_{1:t-2}, u_{1:t-1}})\\
=&p_\theta(h_{1:t-1},r_{1:t-1}|u_{1:t-1})p_\theta(h_t,r_t|h_{t-1}, r_{t-1},u_t)
\end{align*}
It can be seen that in simplifying the above equations, those conditional independence properties hold for our generative model Eq. \eqref{eq:LS-2}.
Then,
\begin{align*}
&p_\theta(h_{1:t}|u_{1:t},r_{1:t})=\frac{p_\theta(h_{1:t},r_{1:t}|u_{1:t})}{p_\theta(r_{1:t}|u_{1:t})}\\
=&\frac{p_\theta(h_{1:t-1},r_{1:t-1}|u_{1:t-1})p_\theta(h_t,r_t|h_{t-1}, r_{t-1},u_t)}{p_\theta(r_{1:t}|u_{1:t})}\\
=&p_\theta(h_{1:t-1}|u_{1:t-1},r_{1:t-1})p_\theta(h_t,r_t|h_{t-1}, r_{t-1},u_t)\\
& \times \frac{p_\theta(r_{1:t-1}|u_{1:t-1})}{p_\theta(r_{1:t}|u_{1:t})}\\
\propto &p_\theta(h_{1:t-1}|u_{1:t-1},r_{1:t-1})p_\theta(h_t,r_t|h_{t-1}, r_{t-1},u_t)
\end{align*}

\section{Implementation Details}
\label{sec:implementation}
We implement the models with Huggingface Transformers repository of version 4.8.2. We initialize both the generative model and the inference model with DistilGPT-2, a distilled version of GPT2. For all of supervised pre-training, variational learning and JSA learning, we use the AdamW optimizer and a linear scheduler with 20\% warm up steps and maximum learning rate $10^{-4}$. The minibatch base size is set to be 8 with gradient accumulation steps of 4. The 3 random seeds for the results in Table \ref{compare-results} are 9, 10 and 11. The total epochs for supervised pre-training are 50, and those for both variational learning and JSA learning are 40. We monitor the performance on the validation set and apply early stopping (stop when the current best model is not exceeded by models in the following 4 epochs). We select the best model on the validation set, then evaluate it on test set. All our experiments are performed on a single 32GB Tesla-V100 GPU.

\end{document}